\documentclass[conference]{IEEEtran}
\IEEEoverridecommandlockouts

\usepackage{cite}
\usepackage{amsmath,amssymb,amsfonts}
\usepackage{algorithmic}
\usepackage{graphicx}
\usepackage{textcomp}
\usepackage{xcolor}
\usepackage{booktabs}
\usepackage{subcaption}
\usepackage{bbm, dsfont}
\usepackage{hyperref}

\def\BibTeX{{\rm B\kern-.05em{\sc i\kern-.025em b}\kern-.08em
    T\kern-.1667em\lower.7ex\hbox{E}\kern-.125emX}}
\begin{document}

\title{Efficient Hierarchical Contrastive Self-supervising Learning for Time Series Classification via Importance-aware Resolution Selection}

\author{
    \IEEEauthorblockN{Kevin Garcia \IEEEauthorrefmark{1}, Juan M Perez\IEEEauthorrefmark{1}, Yifeng Gao\IEEEauthorrefmark{1}}
    \IEEEauthorblockA{\IEEEauthorrefmark{1}Department of Computer Science, 
     University of Texas Rio Grande Valley, 
     \\
     \{kevin.garcia09, juan.m.perez02, yifeng.gao\}@utrgv.edu}
}

\maketitle

\begin{abstract}.
Recently, there has been a significant advancement in designing Self-Supervised Learning (SSL) frameworks for time series data to reduce the dependency on data labels. Among these works, hierarchical contrastive learning-based SSL frameworks, which learn representations by contrasting data embeddings at multiple resolutions, have gained considerable attention. Due to their ability to gather more information, they exhibit better generalization in various downstream tasks. However, when the time series data length is significant long, the computational cost is often significantly higher than that of other SSL frameworks. In this paper, to address this challenge, we propose an efficient way to train hierarchical contrastive learning models. Inspired by the fact that each resolution's data embedding is highly dependent, we introduce importance-aware resolution selection based training framework to reduce the computational cost. In the experiment, we demonstrate that the proposed method significantly improves training time while preserving the original model's integrity in extensive time series classification performance evaluations. Our code could be found here: \url{https://github.com/KEEBVIN/IARS}

\end{abstract}

\begin{IEEEkeywords}
Time series, Self Supervised Learning, Data Mining, Machine Learning
\end{IEEEkeywords}

\section{Introduction}

Time series data is a crucial form of information that has vast applications ~\cite{gensler2016deep,sharadga2020time,wang2013bag,kampouraki2008heartbeat,baisch1999spectral,takanami1991estimation,varotsos2011natural}. Recently, with the widespread use of sensor networks, large-scale time series data has become ubiquitous. Such data gives us a dense amount of valuable information. The task of mining time series could help us harvest important trends, patterns, and crucial behaviors, which ultimately benefits various applications.

A well-known bottleneck in time series related applications is the lack of human labeling\cite{esling2012time}. Typically, only proficient domain experts have the capability to effectively label time series. Such data scarcity issues greatly impact the development of deep learning based approaches since the training process often highly relies on large-scale annotated data samples. To address the challenges above, attention towards Self-Supervised Learning (SSL) based training frameworks has been growing in popularity. These general training frameworks aim to learn a deep learning model that could transform time series data into low-dimensional representations that can represent their semantic information and can be fine-tuned to be suitable for various downstream tasks\cite{yue2022ts2vec} without heavily relying on human supervision. Different from self-supervised learning frameworks (SSL) for image, video, and natural language representation learning, ~\cite{zhai2019s4l,misra2020self,hendrycks2019using,chen2020simple,he2020momentum,chen2020improved,grill2020bootstrap} which has achieved a great success, SSL frameworks for time series data often requires uniquely designed training frameworks to capture multi-resolution semantic information that exist in the time series. For example, in one of the current state-of-the-art SSL frameworks in time series is called ts2vec \cite{yue2022ts2vec}, a multi-resolution hierarchical contrastive learning framework is used for training the model and captures the semantic information in different fidelity, which increases the generalization ability of the learnt representation against various benchmarks. Similarly, observations have also been seen in other works that highlight the importance of such multi-resolution hierarchical contrastive learning schema\cite{wang2024contrast,trirat2024universal}. 

While hierarchical contrastive learning outperforms existing state-of-the-art models, the model's computational cost is much heavier than other self-supervised learning frameworks\cite{chen2020simple}, especially when the time series is much longer in length. This is due to the process of capturing multi-resolution information. The hierarchical enumeration to compute the loss in each resolution of the time series significantly increases the computational burden. In this work, we point out that to capture such mutli-resolution semantic information, an exhausted enumeration is not the only option. Due to the natural correlation across different resolutions, optimizing a training loss in a single resolution can indeed lead to the optimization of neighboring resolutions. Based on this observation, we propose an Importance-aware resolution selection based on a hierarchical contrastive training framework. In each training epoch, we propose an importance aware resolution selection strategy to adaptively select the most important resolutions throughout the training and only optimize a \textbf{single resolution} in each epoch. Such a training process can significantly reduce the computational cost while maintaining similiar generalization ability. We conduct extensive experiments in the task of time series classification to justify our claim. Additionally, the efficiency of the proposed framework is significantly enhanced, notably in large-scale datasets. 

In summary, the contribution of the proposed work is:

\begin{itemize}
    \item We proposed a novel Importance-aware resolution selection based SSL training framework for time series representation learning. During the training, the method can adaptively select the most important resolution to optimize, based on the evolving loss value.  
    
    \item The proposed framework can achieve significant improvements in terms of efficiency while maintaining the downstream task's  performance in generalization.
    
\end{itemize}

The rest of the paper is organized as the following: Section II will discuss the recent advancements in the field of self-supervised learning research. We then describe the problem setting in Section III. Our proposed method is introduced in Section IV. Experimentation is then introduced in Section V. Finally, we conclude this paper in Section VI.

\section{Related Work}

Recently, the self-supervised learning (SSL) framework~\cite{zhai2019s4l,misra2020self,hendrycks2019using,chen2020simple,he2020momentum,chen2020improved,grill2020bootstrap} is introduced for the research domain of computer vision. The goal of SSL is to train a deep learning model to understand the semantic-level invariance characteristics through carefully designed pretext tasks from  high-level semantic understanding related to image data (e.g. learning rotation-invariant representation for images)~\cite{zhai2019s4l,jang2018grasp2vec,gidaris2018unsupervised}. An increasingly amount of time series representation learning research has been focused on designing the self-supervised deep learning framework~\cite{yue2022ts2vec,franceschi2019unsupervised,tonekaboni2021unsupervised,eldele2021time}. Most models are designed based on the unsupervised contrastive learning framework such as SimCLR ~\cite{chen2020simple}. 

Franceschi et al.\cite{franceschi2019unsupervised} introduces an unsupervised contrastive learning framework by introducing a novel triplet selection approach based on a segment's context. Similarly, Tonekaboni et al. proposed a framework named Temporal Neighborhood Coding (TNC) \cite{tonekaboni2021unsupervised}. TNC aims to utilize the temporal correlation along neighboring segments to learn representations. Eldele et al. \cite{eldele2021time} introduces a Temporal and Contextual Contrast (TS-TCC) based framework.  In TS-TCC, two types of augments, strong augmentation and weak augmentation are used to perform contrastive learning. Zhang et al. \cite{zhang2022self} proposed a time-frequency consistent loss (TFC) for contrastive learning where temporal and frequency of the same neighborhoods are pushed closer together often focus on the problem of out-of-distribution generalization ability. 

Recently, Yue et al. \cite{yue2022ts2vec} proposed a framework named ts2vec. The proposed framework introduces a random cropping based augmentation and a hierarchical loss, which aims to extract semantic information in all resolutions, in order to obtain a generalizeable embedding representation. Wang et al.\cite{wang2024contrast} proposed a task-driven multi-resolution contrastive learning framework, which aims at contrasting semantic information in different levels for medicial time series.  Both achieved significantly better performance compared with previous methods. However, the extensive computational burden of performing contrastive loss in all resolutions also leads to higher comptuational cost.

\section{Problem Setting}

We next describe the definitions and problem statement of hierarchical time series contrastive self-supervised learning \cite{yue2022ts2vec}.

Suppose a time series $T=t_1,\dots,t_L$ is a set of observations ordered by time where $t_i \in R^M$. Given $N$ time series data $\mathcal{D}=\{T_1, T_2,\dots, T_N\}$, the goal of the self-supervised learning is to find a non-linear mapping function $h(\cdot): T_i \rightarrow F_i$ which maps all the \textit{raw time series data} $T_i$ into a $K\times L$-dimensional latent space feature map $F_i$ such that the semantic information of all levels are preserved without the help from any human annotation. In the fine-tunning stage, the learnt feature map $e_i$ has the ability to directly apply to down-steam tasks.

Note that unlike a traditional SSL framework which focuses on learning an embedding vector of $K$ dimensions for downstream tasks, in hierarchical time series contrastive self-supervised learning\cite{yue2022ts2vec}, the framework aims to learn representation feature maps $F_i$ instead. The representation feature can effectively generate different resolution's embedding vector by global average pooling preserving semantic information over dynamic time spans, and has a higher ability to transform  knowledge to downstream tasks\cite{yue2022ts2vec,wang2024contrast}.

\section{Methodology}

In this section, we will first introduce the overall framework. We will then introduce the proposed importance aware resolution selection training method. Intuitively, the framework aims to choose the "most important" resolution to train, instead of training the loss of every resolution in every epoch, which enhances the training time while maintaining the same model performance compared to the original aggregation approach.

\begin{figure}[h]
    \centering
    \includegraphics[scale=0.54]{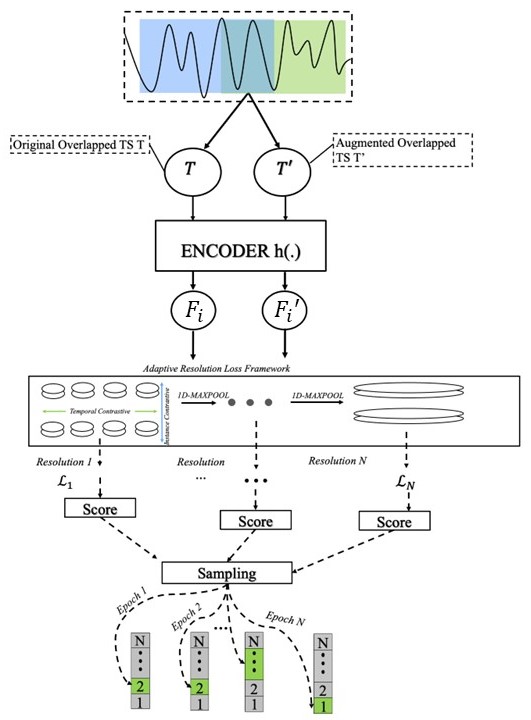}
    \caption{Proposed Framework: Starting with original and augmented time series \textbf{$T$} and \textbf{$T'$}, the data is encoded into representations \textbf{$z$} and \textbf{$z'$}. The framework computes loss across multiple resolutions using Temporal Contrastive methods combined with 1-Dimensional MaxPooling. The framework calculates losses \boldmath $\mathcal{L}_1$ to \boldmath $\mathcal{L}_N$ across multiple resolutions. Each loss is associated with a score that evaluates its significance. Based on these scores, the importance-aware sampling decides which resolution's loss to focus on for a specific training epoch.}
\end{figure}

\subsection{Overall Framework}
The current framework of the hierarchical contrastive loss function computes an aggregation of multiple resolutions throughout training, which yields a high computational load. Our approach eliminates this high computational cost via a scoring and sampling system. The score function assigns an importance value to each resolution which is based on the current loss value of the respective resolution. A sampling is then performed which is based on the respective score values, and the training resumes on the current 'most important' resolution.

Our proposed framework is illustrated in Figure 1. The overall framework consists of three components. First, a augmentation process to augment semantically similar samples, an encoder backbone to obtain pre-trained feature maps, and an importance-aware resolution selection based loss to train the model. Specifically, given $B$ number of $D$-dimensional multivariate time series data of length $N$ ($T\in R^{B\times D \times N}$), in the augumentation step, we perform a cropping and time-step masking based operation to generate two different views of the data\cite{yue2022ts2vec}. Each view will be considered as the data that shared similar semantic meaning and will be used to train the learnable encoder model that can recognize such meaning. Next, a learnable encoder module $h(.)$ is applied to extract the semantic information represented in latent embedding features ($F_i$ and $F'_i$, respectively for two different views). Lastly, the framework will calculate the importance of each resolution based on score representations and adaptively select a single resolution to optimize via a probabilistic sampling. For experimentation, the encoder $h(.)$ is modeled via a 1-D convolutional layer based ResNet18 model for feature extraction. 

\subsection{Augmentation}

\subsubsection{Cropping based View Augmentation}

Given an input time series, the model first generates two augmentations based on random cropping. In this step, the model generates two overlapped sub-sequences $T$ and $T'$ where $T\cap T' \neq \emptyset$. $T$ and $T'$ will be used to train the encoder $h(.)$.The cropping operation generates two subsequences that contain a shared region. Such shared regions should have much similar representations compared with other random regions. Thus, the framework aims to pretrain the encoder to recognize and ensure the feature maps represent an \textbf{overlapped} region between $T$ and $T'$ ($h(T)\cap h(T')$). They should be similar to each other in comparison to other randomly sampled subsequences.

\subsubsection{Time Stamp Masking Module} 
Time series data samples are highly correlated. Therefore, the semantic information of the data will not change if we mask random time steps. In the embedding feature map, a random masking is applied to generate an augmented context view by masking latent vectors at randomly selected timestamps (via dropout). It essentially hides some of the information by creating a slightly different version of the data from the original, allowing the model to learn more robust representations. 

\subsection{Representation Learning}

Given two time series views $T \in R^{M\times D \times n}$ and $T' \in R^{M\times D \times n}$. Both the cropped-and-masked subsequences will pass through the encoder to obtain the embeddings $F=h(T)$ and $F'=h(T')$, where $F \in R^{M\times D' \times n}$ and $F' \in R^{M\times D' \times n'}$). While the arbitrary encoder function $h(.)$ can be used in this step, we will use a 1-D ResNet18 backbone \cite{he2016deep,ismail2019deep} without final pooling and FCN layers. Unlike the original ResNet18 framework, the used backbone replaces all the 2-D convolution operators to 1-D convolution:
$y = x + Conv1D(x)$. It has been shown that such ResNet backbones can achieve reasonable performance in various time series data mining tasks \cite{ismail2019deep}.

\subsection{Proposed Contrastive Learning Loss}

\subsubsection{Instance and Temporal Contrastive Loss}
After two feature maps $F$ and $F'$ are obtained through the encoder, a contrastive learning training loss is computed at each resolution $r$. Given $F$ and $F'$, the contrastive learning loss is forced to push the latent representations of the \textit{overlapping region} of $T$ and $T'$ to be similar. Specifically, the latent  sequence $F$ and $F'$ are first down-sampled to resolution $r$. In this step, $F_{o}$ and $F'_{o}$ are all down sampled to a sequence only consisting of $r$ time stamps via average pooling:

\begin{equation}
    F_{r} = AvgPool(F_{o}, r) 
\end{equation}

\begin{equation}
    F'_{r} = AvgPool(F'_{o}, r) 
\end{equation}

and the overlapping region feature map is denoted as:

\begin{equation}
    F_{o,r} = \{f_{:,:,k}|k \in F_{r}\cap F'_{r}\}
\end{equation}

\begin{equation}
    F'_{o,r} = \{f'_{:,:,k}|k \in F_{r}\cap F'_{r}\}
\end{equation}

The framework performs two types of contrastive losses: 

\textbf{temporal-wise contrastive loss}: The idea is illustrated in Figure 2, intuitively, the loss considers the aligned timestamps belonging to $F_{o,r}$ and $F'_{o,r}$ as positive samples (highlighted in green arrows) and the rest unrelated time steps as the negative samples (highlighted in red arrows). The loss can be written as:

\begin{figure}[!h]
    \centering
    \includegraphics[scale=0.13]{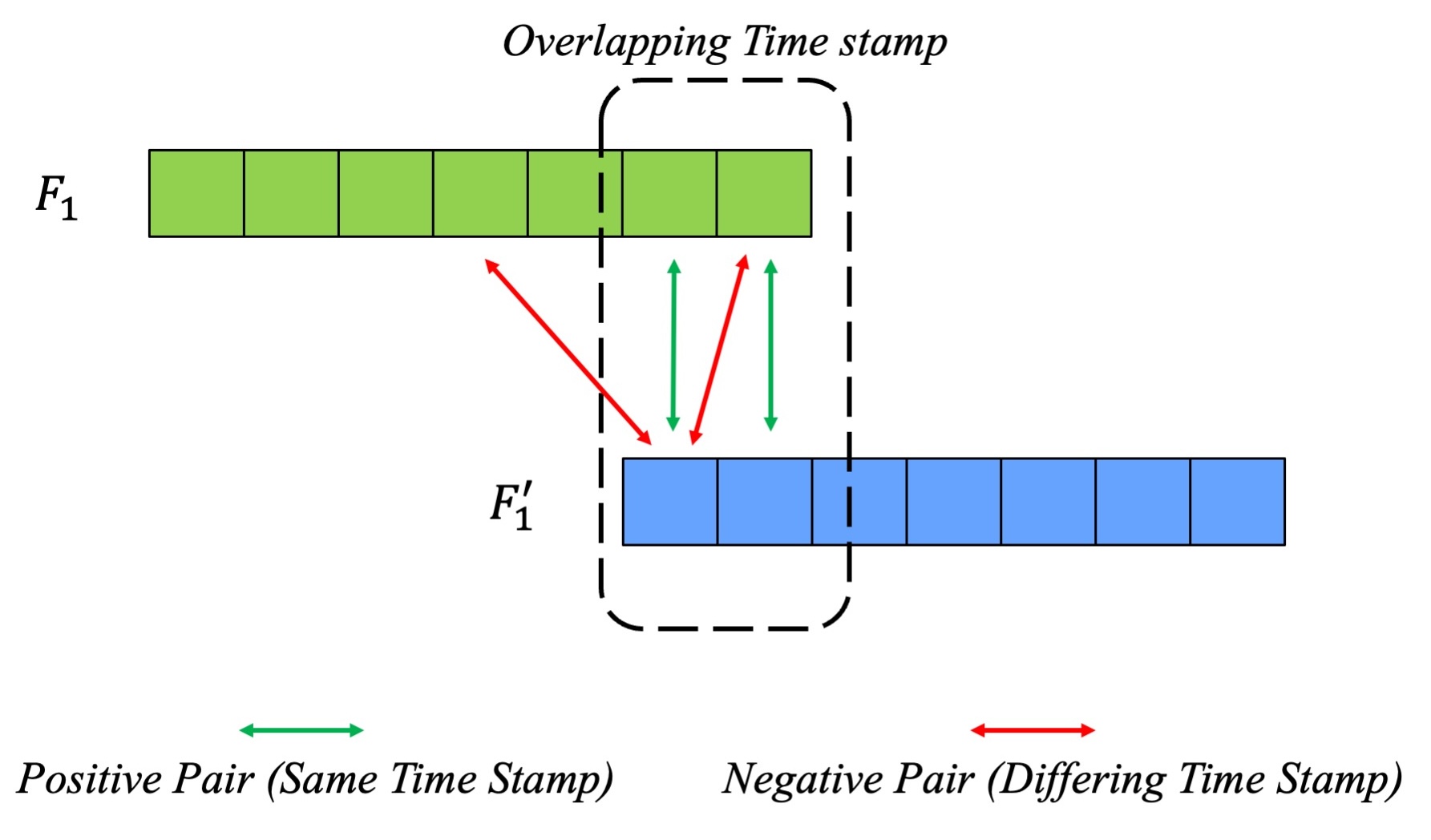}
    \caption{Temporal Wise Loss}
\end{figure}

\begin{equation}
    \resizebox{0.85\hsize}{!}{$
    \ell^{temp}_r = \sum_{i} \sum_{t} -log{\frac{exp{( f^{(o)}_{i,t} \cdot f^{o'}_{i,t})}}{\sum_{t'} \Bigl( exp{( f_{i,t} \cdot f'_{i,t'})} +\mathbbm{1}_{[t \ne t']} exp{( f_{i,t} \cdot f_{i,t'})}\Bigl) } } 
    $}
\end{equation}

where $f^{(o)}_{i,t} \in  F_{o}$ and $f^{o'}_{i,t} \in  F_{o'}$ represents the feature map of the overlapping region. $f_{i,t} \in  F$ and $f'_{i,t} \in  F'$ represents the entire feature map generated from $T$ and $T'$, respectively.

\begin{figure}[!h]
    \centering
    \includegraphics[scale=0.13]{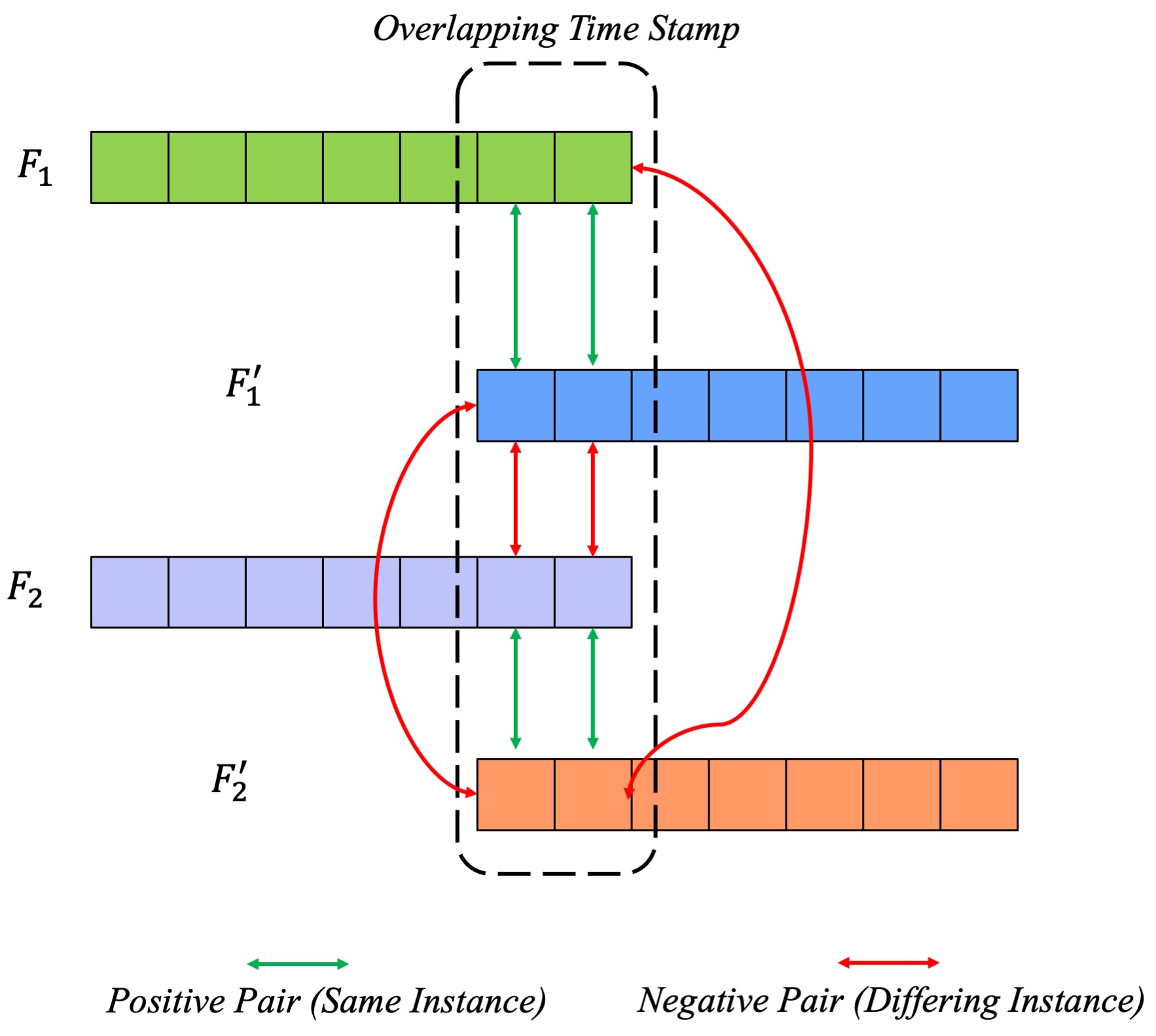}
    \caption{Instance Wise Loss}
\end{figure}

\textbf{instance-wise contrastive loss}: The idea is illustrated in Figure 3, which considers the feature map values in other time series instances as negative samples:

\begin{figure*}[t]
 \centering
 \includegraphics[width=150mm]{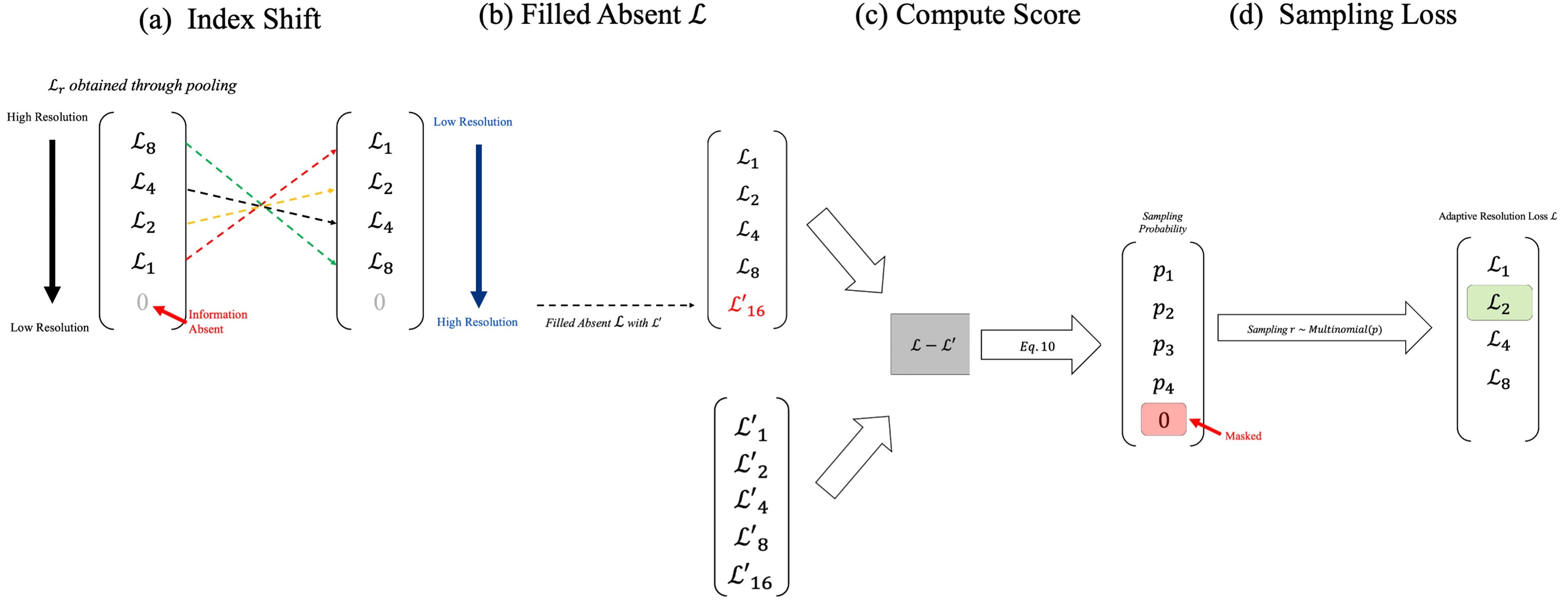}
\caption{Importance-aware Resolution Selection Overview: \textbf{(a)}: aligned resolution loss $\mathcal{L}$. \textbf{(b)}: Every absented loss value is replaced by its prior recorded loss $\mathcal{L}'$. \textbf{(c)}: Each loss is passed through Equation 10 and generates a probability. \textbf{(d)}:Sampling from the generated distribution.}
\end{figure*}

\begin{equation}
\resizebox{0.9\hsize}{!}{$
    \ell^{inst}_r = \sum_{i} \sum_{t}  -log{\frac{exp{( f^{(o)}_{i,t} \cdot f^{o'}_{i,t})}}{\sum_{j=1}^B \Bigl(exp{( f_{i,t} \cdot f'_{j,t})} + \mathbbm{1}_{[i \ne j]} exp{( f_{i,t} \cdot f_{j,t})}\Bigl) } }
   $}
\end{equation}

Afterwards, both losses are then added together to create an overall loss value:

\begin{equation}
    \mathcal{L}_r=\alpha \ell^{temp}_r + (1-\alpha) \ell^{inst}_r
\end{equation}

where $\alpha$ is a hyper parameter that controls the importance of temporal and instance losses.

Finally, the hierarchical contrastive loss is defined as

 \begin{equation}
     \mathcal{L}^{hier} = \sum_{i=1,i<\log m}\mathcal{L}_{2^i}
 \end{equation}. 

 where $m$ denotes the resolution, which is used to capture the semantic similarity of time series in multiple resolutions.

\subsubsection{Importance Aware Resolution Selection}

In this subsection, we will introduce our proposed importance aware resolution selection. We measure the \textbf{loss evolving trend}, as an indicator of whether the loss is indirectly co-trained in a specific epoch when optimizing the loss of another resolution, and we then utilize it to determine which $\mathcal{L}_r$ will be picked in the next epoch. The overall framework is shown in Figure 4. It consists of four different steps: 1) an index shifting step to align resolution indexes, 2) an absent value filling step to ensure the consistency of observed resolutions in different epochs, 3) a ranking step to compute the score for each resolution, and 4) lastly, a probabilistic sampling step to select a resolution based on the ranking score.

\textbf{Preprocessing}: Since the overlapping area between $T$ and $T'$ can be of an arbitrary length between $[1,L-1]$, the resolution indices $r$ in different epochs do not represent the same resolution. Therefore, before computing the actual importance score, the algorithm will first align each resolution across different epochs (Fig. 4 (a-b)). Specifically, the algorithm reorders the resolutions and obtains the number of resolutions from the given dataset and interpolates the most current loss value for any absent resolution. As the model trains, some resolution's losses may be missing due to the sampling length. Our process reuses the previous non-zero loss to mitigate the issue of having a loss value in an absent resolution.

\textbf{Importance Score}: Throughout the training, the losses trend downward when $\mathcal{L}_r$ is indirectly trained and the loss either plateaus or increases when $\mathcal{L}_r$ cannot be indirectly trained. Knowing this, proceeding the index shift and filled absent steps, a score function is then used to evaluate the importance of each resolution. 

Specifically, given the loss value of $\mathcal{L}_{r',e}$ in the $e_{th}$ epoch of resolution $i$, we measure this co-training behavior based on: 

\begin{align}
s_{i} = \frac{\exp(\mathcal{L}_{i,e}   - \mathcal{L}_{i,{e-1}} )}{\sum_{r \in \mathcal{D}} \exp(\mathcal{L}_{r,e}   - \mathcal{L}_{r,{e-1}}) }
\end{align}

where $\mathcal{D}$ denotes all resolutions used to compute $\mathcal{L}^{hier}$. Equation 7 measures whether the loss can be indirectly trained and normalizes the value scale.

The algorithm will compute the score via a softmax function based on the respective resolution's loss to normalize the score between 0 and 1. 

\begin{equation}
    \bar{s} = softmax(s)
\end{equation}

The score is correlated to the loss value given, the greater the loss the greater the score assigned. This generates a probability array that we can then use to select the most important resolution based on a random multinomial function. Note that this function ignores any absented resolution in the original loss array, which means this array is usually smaller than $\log L$ size.

\textbf{Probabilistic Sampling}: In this step, we pick the loss value that cannot be indirectly trained in the current epoch to optimize (shown in Fig. 4(c)). We select the resolution with the relatively worst performance. The poorest performing loss value gives us a clear representation of the current performance and indicates the needs of optimizing this resolution. Specifically, the algorithm will sample one resolution $r'$ from a multinomial distribution determined by the important score $s$: 

\begin{equation}
    r' \sim Multinomial(s)
\end{equation}

Once $r'$ is obtained, the contrastive loss obtained at resolution $r'$ is then used to update the model's weights, giving the chosen resolution priority over the rest:

\begin{equation}
     \mathcal{L}^{a} =  \mathcal{L}_{r'}
\end{equation}
  
Overall, the model trained on $\mathcal{L}^{a}$ will emphasize on selecting the "worst" resolution to perform the optimization, which increases the efficiency of training the SSL model without performing gradient descent over all selected resolutions. The back-propagation computation will perform only on a single loss $\mathcal{L}^{a}$, which greatly decreases the computational burden.

\section{Experimentation}

\subsection{Experiment Setup}

In this section, we will comprehensively analyze the performance of the proposed method. We evaluate the proposed model on time series classification task \cite{yue2022ts2vec}. We use all UEA/UCR multi-variate time series datasets that are of lengths between 400 to 3000 to evaluate the performance. All of the experiments are conducted in Google Colab with an NVIDIA T4 GPU on most of the data and an A100 GPU was used if the T4 cannot fulfill the computational resources. We evaluate the effectiveness via classification accuracy and evaluate the efficiency by the training time. All comparison experiments were repeated five times for each dataset and the average performance was recorded. The information of the selected datasets are shown in Table I.

Throughout the experiments in this paper, the hyper-parameters are fixed across all the experiments unless otherwise noted. The learning rate is set to be $1e-3$, and the embedding dimension is set to 128. The final embedding is computed through global average pooling across all time stamps during each comparison evaluation. All the experiments have been done three times and reports the average performance. Each subsection has additional information regarding the specific parameter changes that occurred. 

\begin{table}[!h]
\centering
\resizebox{\columnwidth}{!}{%

\begin{tabular}{@{}lcccccc@{}}
 
\toprule
Dataset              & Abbrev.    & Train Size & Test Size & Length & No. of Classes & Type   \\ \midrule
HandMovementDirection & HMD & 160        & 74        & 400    & 4              & EEG    \\ \midrule
HeartBeat            & HB & 204        & 205       & 405    & 2              & AUDIO  \\ \midrule
AtrialFibrilation   & AF  & 15         & 15        & 640    & 3              & ECG    \\ \midrule
SelfRegulationSCP1   & SCP1 & 268        & 293       & 896    & 2              & EEG    \\ \midrule
PhoneMe             & PM  & 214        & 1896      & 1024   & 39             & SOUND  \\ \midrule
SelfRegulationSCP2   & SCP2 & 200        & 180       & 1152   & 2              & EEG    \\ \midrule
Cricket            & Cr   & 108        & 72        & 1197   & 12             & HAR    \\ \midrule
EthanolConcentration  & EC & 261        & 263       & 1751   & 4              & OTHER  \\ \midrule
StandWalkJump      & SWJ   & 12         & 15        & 2500   & 3              & ECG    \\ \midrule
MotorImagery     & MI     & 278        & 100       & 3000   & 2              & EEG    \\ \bottomrule
\end{tabular}%
}
\caption{Tested Datasets Information}
\end{table}

\subsection{Datasets}
Each Dataset's characteristics that were used for experimentation are shown in Table I. We opted for datasets with a larger length, as our observations indicated that a larger dataset length could potentially result in a substantial decrease in training time. Each dataset is used for the downstream task of multivariate time series classification, and are standard baseline datasets used across various applications and research areas. Each dataset varies by training size, testing size, length, number of classes and type. This shows that our proposed implementation is robust regardless of dataset type and/or characteristics. The abbreviation's meaning and each dataset's characteristics are shown as follows:

\subsubsection{HMD} This dataset is recorded through EEG, on two right handed patients using a joystick for audio prompts. It has a sampling rate of 400 Hz, and recording interval of -0.4s (before movement) and 0.6s (after movement) totaling a 1 second window (excluding the movement time).

\subsubsection{HB} This dataset is recorded via audio, and has 2 types of participants. Those with normal and abnormal heartbeats, the participants with abnormal heartbeats are diagnosed with a cardiac diagnosis. Each recording is 5 seconds, and there are 113 and 296 samples of normal and abnormal heart beats respectively.

\subsubsection{AF} This dataset is composed of 5s recordings of ECG data with a sample rate of 128 per second, with each sample having 2 ECG signals. It has 3 classes determining 3 levels of atrial fibrilation, non-terminal, self-terminating and immediate termination.

\subsubsection{SCP1} This dataset is recorded via EEG on participants visualizing moving a cursor up or down a screen. The recording was done 6 times from different positions on the head. It is used for the downstream task of classification.

\subsubsection{PM} This dataset is a collection of audio files turned into spectrograms that were sampled at 22,050 Hz from male and female speakers. Each instance has a window size of 0.001s and have an overlap of 90\%. There are 39 classes total and each have 170 instances.  

\subsubsection{SCP2} This dataset is recorded via EEG on artificially respirated ALS patients with the same task as those from SCP1. The recording was done 7 times from different positions on the head. It is used for the downstream task of classification.

\subsubsection{Cr} This dataset records hand gestures via 2 accelerometers at a frequency of 184Hz, 12 gestures are repeated 10 times on 3 axes per accelerometer x, y and z. 

\subsubsection{EC} This dataset is recorded as a spectra of water-ethanol solution, with 4 different classes which represent ethanol content in the solution. The data is organized as follows, each instance has 3 repeated recordings of the same bottle and batch of solution. 

\subsubsection{SWJ} This data is recorded via ECG sensors placed on a male participant standing, walking and jumping (3 classes). A spectrogram was produced from the original data with a window size of 0.006 seconds with an overlap of 70\%. Each instance is a frequency band from the spectrogram. Each class has 9 instances. 

\subsubsection{MI} This dataset is recorded via EEG data, and contains 64 dimensions. Where the participant is imagining movement on their tongue or finger. Each recording has a length of 3 seconds.

\subsection{Baselines}

In the experimentation, we first compare the proposed method with ts2vec, the state-of-the-art SSL training framework that trained with hierarchical contrastive loss. Then, we will evaluate the proposed method with classical 1-NN based time series classification baselines. In summary, the baselines used in the experiment can be summarized in the following:

\subsubsection{ts2vec ($\mathcal{L}^{hier})$} The state-of-the-art SSL training framework that is trained based on the hierarchical contrastive loss $\mathcal{L}^{hier})$.

\begin{table}[t]
\centering
\resizebox{0.85\columnwidth}{!}{%
\begin{tabular}{@{}lcccc@{}}
\toprule
\multicolumn{5}{c}{K = 16}                                                                              \\ \midrule
\multicolumn{1}{c}{}        & \multicolumn{2}{c}{$\mathcal{L}^{hier}$ }        & \multicolumn{2}{c}{$\mathcal{L}^{a}$ }        \\ \midrule
Dataset                     & Execution Time (s) & Accuracy       & Execution Time (s) & Accuracy       \\ \midrule
HandMovementDirection       & 128.40             & 0.286          & \textbf{78.19}     & \textbf{0.300} \\ \midrule
HeartBeat                   & 175.07             & \textbf{0.738} & \textbf{98.36}     & 0.730          \\ \midrule
AtrialFibrilation           & 81.37              & 0.267          & \textbf{67.17}     & \textbf{0.293} \\ \midrule
SelfRegulationSCP1          & 187.74             & 0.808          & \textbf{122.39}    & \textbf{0.814} \\ \midrule
Phoneme                     & 585.87             & \textbf{0.148} & \textbf{428.97}    & 0.147          \\ \midrule
SelfRegulationSCP2          & 200.00             & 0.526          & \textbf{129.04}    & \textbf{0.549} \\ \midrule
Cricket                     & 138.13             & \textbf{0.942} & \textbf{104.96}    & 0.914          \\ \midrule
EthanolConcentration        & 721.60             & 0.278          & \textbf{491.91}    & \textbf{0.282} \\ \midrule
StandWalkJump               & 151.67             & \textbf{0.427} & \textbf{117.48}    & 0.400          \\ \midrule
MotorImagery                & 2073.63            & 0.494          & \textbf{1598.05}   & \textbf{0.542} \\ \midrule
\# of Best performing       & 0                  & 4              & \textbf{10}        & \textbf{6}     \\ \midrule
\multicolumn{5}{c}{K = 32}                                                                              \\ \midrule
\multicolumn{1}{c}{}        & \multicolumn{2}{c}{$\mathcal{L}^{hier}$ }        & \multicolumn{2}{c}{$\mathcal{L}^{a}$ }        \\ \midrule
Dataset                     & Execution Time (s) & Accuracy       & Execution Time (s) & Accuracy       \\ \midrule
HandMovementDirection       & 131.11             & \textbf{0.289} & \textbf{78.57}     & 0.243          \\ \midrule
HeartBeat                   & 180.35             & 0.735          & \textbf{101.06}    & \textbf{0.739} \\ \midrule
AtrialFibrilation           & 70.15              & \textbf{0.360} & \textbf{55.37}     & 0.280          \\ \midrule
SelfRegulationSCP1          & 190.83             & 0.813          & \textbf{122.96}    & \textbf{0.839} \\ \midrule
Phoneme                     & 587.24             & \textbf{0.166} & \textbf{416.13}    & 0.161          \\ \midrule
SelfRegulationSCP2          & 204.99             & 0.512          & \textbf{131.06}    & \textbf{0.536} \\ \midrule
Cricket                     & 140.60             & \textbf{0.939} & \textbf{105.84}    & 0.917          \\ \midrule
EthanolConcentration        & 715.52             & \textbf{0.278} & \textbf{492.64}    & 0.265          \\ \midrule
StandWalkJump               & 152.97             & \textbf{0.413} & \textbf{118.38}    & 0.400          \\ \midrule
MotorImagery                & 2071.34            & \textbf{0.512} & \textbf{1535.67}   & 0.478          \\ \midrule
\# of Best Performing       & 0                  & \textbf{7}     & \textbf{10}        & 3              \\ \midrule
\multicolumn{5}{c}{K = 64}                                                                              \\ \midrule
\multicolumn{1}{c}{}        & \multicolumn{2}{c}{$\mathcal{L}^{hier}$ }        & \multicolumn{2}{c}{$\mathcal{L}^{a}$ }        \\ \midrule
Dataset                     & Execution Time (s) & Accuracy       & Execution Time (s) & Accuracy       \\ \midrule
HandMovementDirection       & 136.17             & \textbf{0.319} & \textbf{80.44}     & 0.268          \\ \midrule
HeartBeat                   & 189.23             & 0.735          & \textbf{106.56}    & \textbf{0.743} \\ \midrule
AtrialFibrilation           & 63.96              & \textbf{0.333} & \textbf{52.35}     & 0.280          \\ \midrule
SelfRegulationSCP1          & 192.16             & 0.837          & \textbf{116.96}    & \textbf{0.840} \\ \midrule
Phoneme                     & 574.80             & \textbf{0.175} & \textbf{389.89}    & 0.167          \\ \midrule
SelfRegulationSCP2          & 207.25             & 0.518          & \textbf{137.07}    & \textbf{0.522} \\ \midrule
Cricket                     & 136.03             & \textbf{0.917} & \textbf{98.07}     & 0.914          \\ \midrule
EthanolConcentration        & 718.50             & \textbf{0.265} & \textbf{486.80}    & \textbf{0.265} \\ \midrule
StandWalkJump               & 142.69             & \textbf{0.44}  & \textbf{109.02}    & 0.400          \\ \midrule
MotorImagery                & 1942.80            & 0.506          & \textbf{1422.25}   & \textbf{0.526} \\ \midrule
\# of Best Performing       & 0                  & \textbf{6}     & \textbf{10}        & 5              \\ \midrule
\multicolumn{5}{c}{K = 128}                                                                             \\ \midrule
\multicolumn{1}{c}{}        & \multicolumn{2}{c}{$\mathcal{L}^{hier}$ }        & \multicolumn{2}{c}{$\mathcal{L}^{a}$ }        \\ \midrule
Dataset                     & Execution Time (s) & Accuracy       & Execution Time (s) & Accuracy       \\ \midrule
HandMovementDirection       & 156.08             & \textbf{0.311} & 96.90              & 0.273          \\ \midrule
HeartBeat                   & 220.25             & \textbf{0.733} & 129.45             & 0.729          \\ \midrule
AtrialFibrilation           & 74.27              & \textbf{0.333} & 62.46              & 0.280          \\ \midrule
SelfRegulationSCP1          & 207.63             & 0.835          & 134.65             & \textbf{0.836} \\ \midrule
Phoneme                     & 619.92             & \textbf{0.165} & 433.51             & 0.162          \\ \midrule
SelfRegulationSCP2          & 231.08             & 0.526          & 130.52             & \textbf{0.547} \\ \midrule
Cricket                     & 146.53             & 0.900          & 110.25             & \textbf{0.928} \\ \midrule
EthanolConcentration        & 763.70             & 0.263          & 533.01             & \textbf{0.272} \\ \midrule
StandWalkJump               & 153.40             & 0.387          & 118.17             & \textbf{0.480} \\ \midrule
MotorImagery                & 2116.81            & 0.492          & 1622.82            & \textbf{0.504} \\ \midrule
\# of Best Performing       & 0                  & 4              & \textbf{10}        & \textbf{6}     \\ \midrule
Total \# of Best Performing & 0                  & \textbf{21}    & \textbf{40}        & 20             \\ \bottomrule
\end{tabular}%
}
\caption{ Embedding Size Experiment: Original Framework v.s. Proposed}
\label{tab:my-table}
\end{table}

\subsubsection{1-NN-DTW-D} A 1-Nearest Neighbor model trained on each dataset using Dynamic Time Warping as its distance metric, with dependency.

\subsubsection{1-NN-DTW-I} A 1-Nearest Neighbor model trained on each dataset using Dynamic Time Warping as its distance metric, with independency.

\subsubsection{1-NN-ED} A 1- nearest neighbor model trained on each dataset using Euclidean Distance as its distance measurement metric.

\subsection{Vs. Original Contrastive Learning Framework}

Next, we evaluate the proposed methods by comparing both accuracy and the training efficiency with the original hierarchical contrastive learning framework. In this subsection, we mainly focus on comparing our approach with the ts2vec model. To comprehensively study the performance improvement, we tested both methods on the classification task with embedding sizes ranging from $16$ to $128$ ($K$ set to be $16$, $32$, $64$, and $128$, respectively). Both training time and the accuracy are reported. The comparison results are shown in Table II. 

In terms of efficiency, the proposed method consistently has a much lower execution time through out all of the experiments. In all experiments, the proposed method is faster than the original framework. The result indicates that the proposed method significantly increases the efficiency of the original model. 

In terms of the accuracy, the proposed method obtains a very similar result compared to the original framework. Among all the 40 testing experiments, our method consistently outperforms the original in $20$ trials, which is similar to the performance of the original ts2vec method ($21$ counts). The results demonstrate that the representation learned through the proposed framework has similar capacity as the ts2vec framework but consistently increases the efficiency. 

Moreover, we note that our implementation considerably outperforms the original model in terms of execution time whilst consistently displaying an insignificant change in accuracy across all embedding sizes. This implies that our proposed implementation is robust regardless of parameter changes. Specifically, with embedding sizes increasing from 16 to 128, there is an insignificant change in execution time. Similar comparison results is observed with accuracy. Across all of the embedding sizes, the method achieves similar performance compared with the original framework. Throughout these trials, we see that we are undoubtedly superior in execution time and note a significant decrease in training time across all trials. In summary, the results show that our model is not only consistent, but it does not jeopardize the model's accuracy. 

A one-to-one comparison illustration is shown in Figure 5. The highlighted region indicates the proposed method achieves better performance. According to the figure, in terms of efficiency, all the dots fall in the highlighted region. Moreover, among the datasets such as MotorImagery which contains 3,000 samples, the method has the most significant improvement in all embedding size settings. This indicates that the proposed method has better performance on long time series. In terms of the accuracy, the proposed method has very similar performance compared with the original framework. Most of the dots are positioned around the diagonal in all embedding size settings, indicating that the performance difference is small in most cases. The result demonstrates that the proposed method has similar generalization ability compared with the original framework.

\begin{figure}
  \begin{subfigure}[t]{.23\textwidth}
    \centering
    \includegraphics[width=\linewidth]{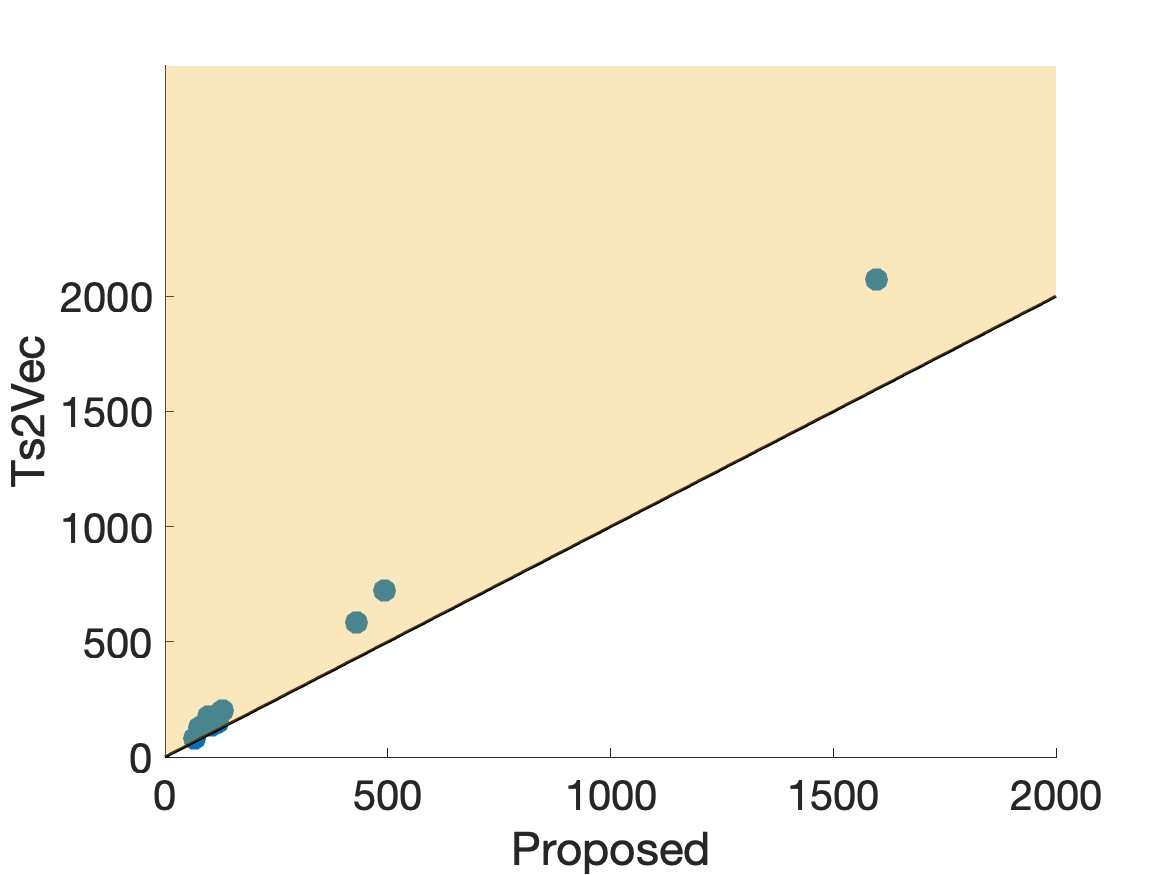}
    \caption{Execution Time ($K=16$)}
  \end{subfigure}
  \hfill
  \begin{subfigure}[t]{.23\textwidth}
    \centering
    \includegraphics[width=\linewidth]{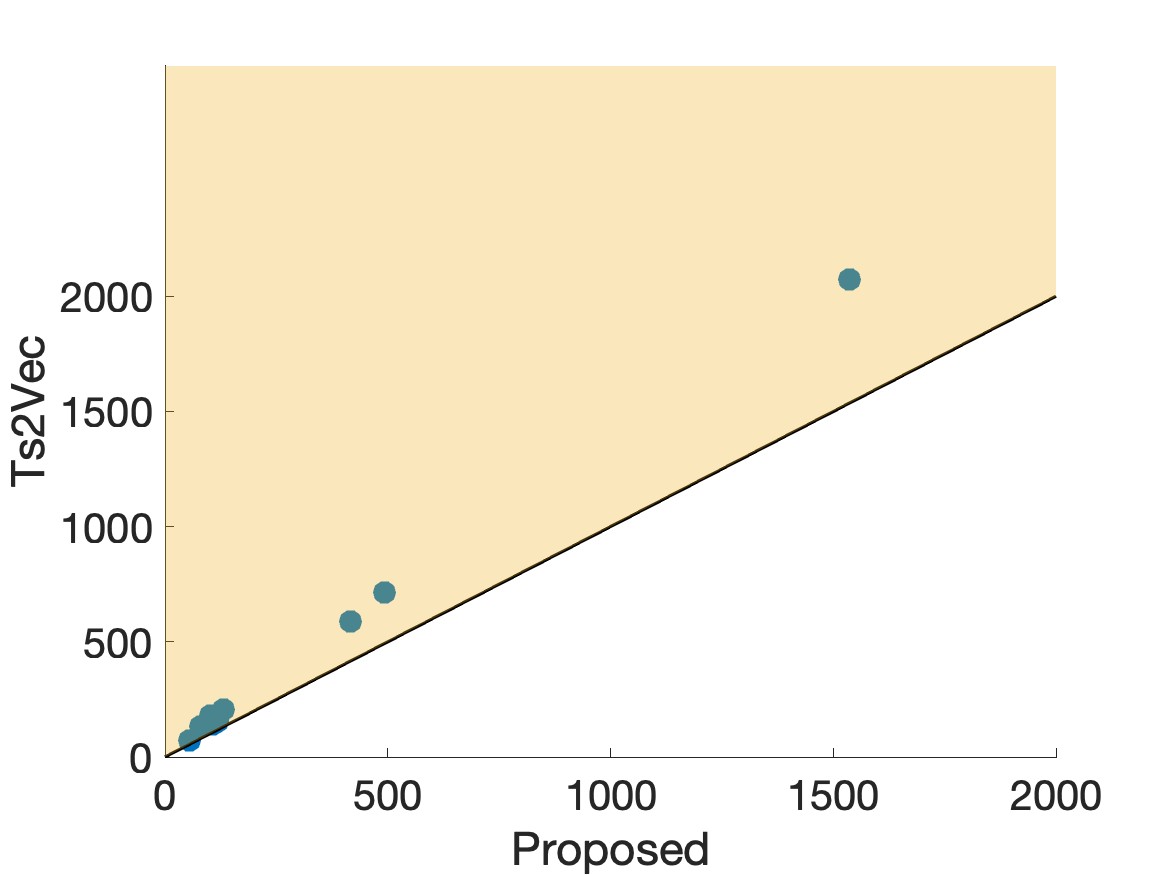}
     \caption{Execution Time ($K=32$)}
  \end{subfigure}

  \medskip

  \begin{subfigure}[t]{.23\textwidth}
    \centering
    \includegraphics[width=\linewidth]{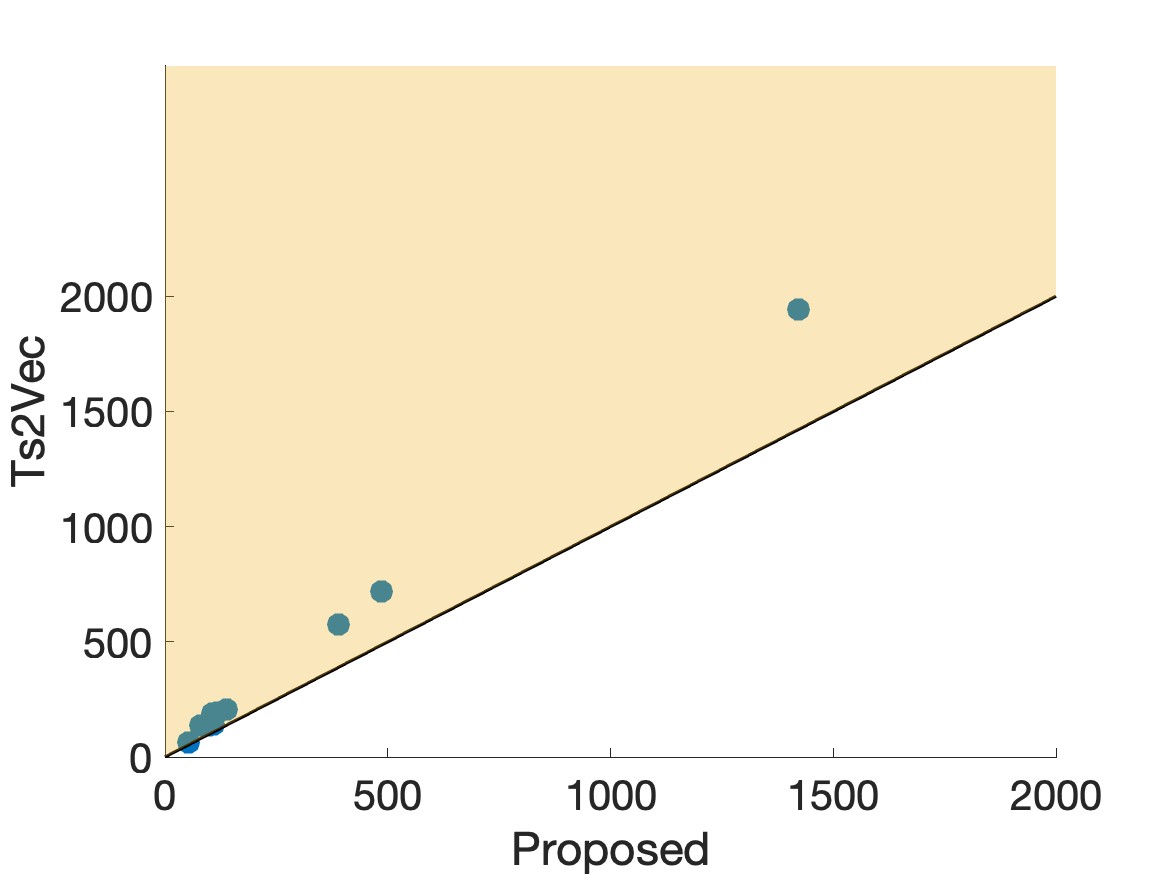}
    \caption{Execution Time ($K=64$)}
  \end{subfigure}
  \hfill
  \begin{subfigure}[t]{.23\textwidth}
    \centering
    \includegraphics[width=\linewidth]{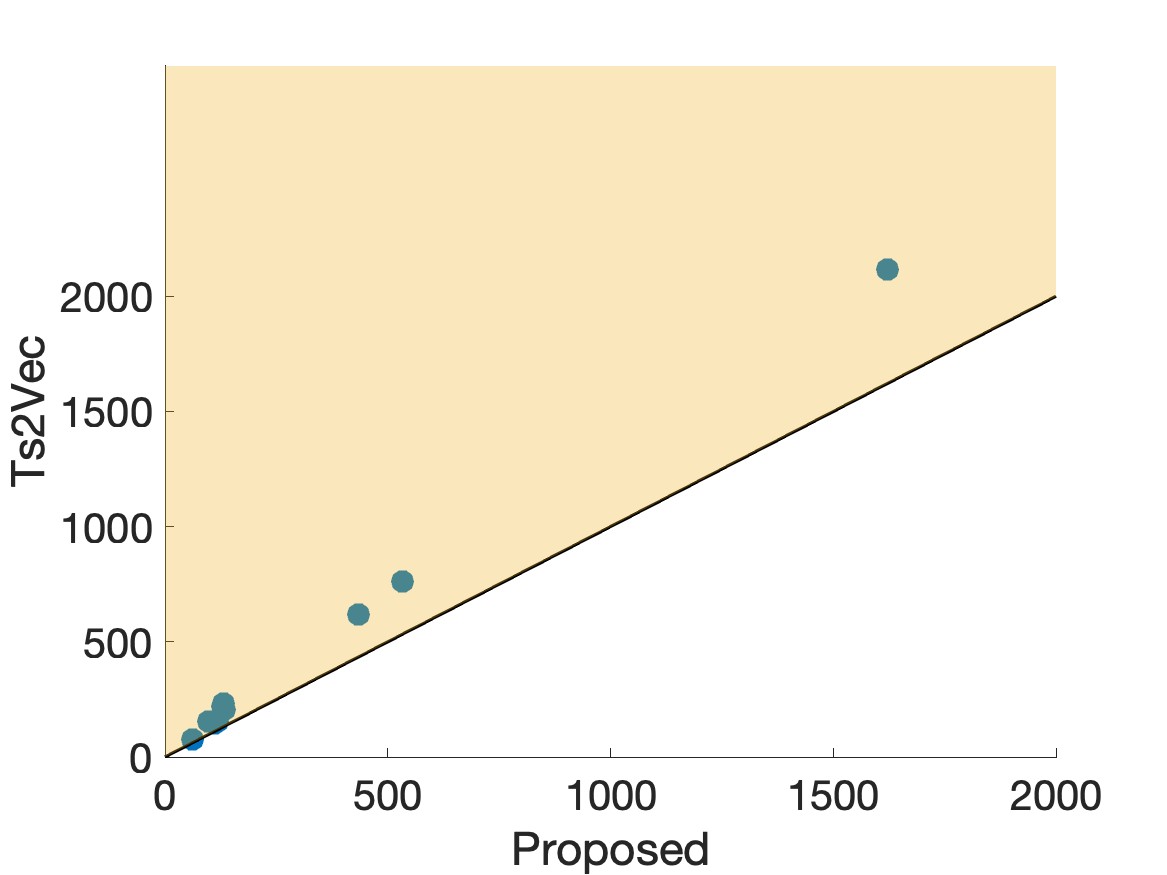}
     \caption{Execution Time ($K=128$)}
  \end{subfigure}

  \medskip
  
  \begin{subfigure}[t]{.23\textwidth}
    \centering
    \includegraphics[width=\linewidth]{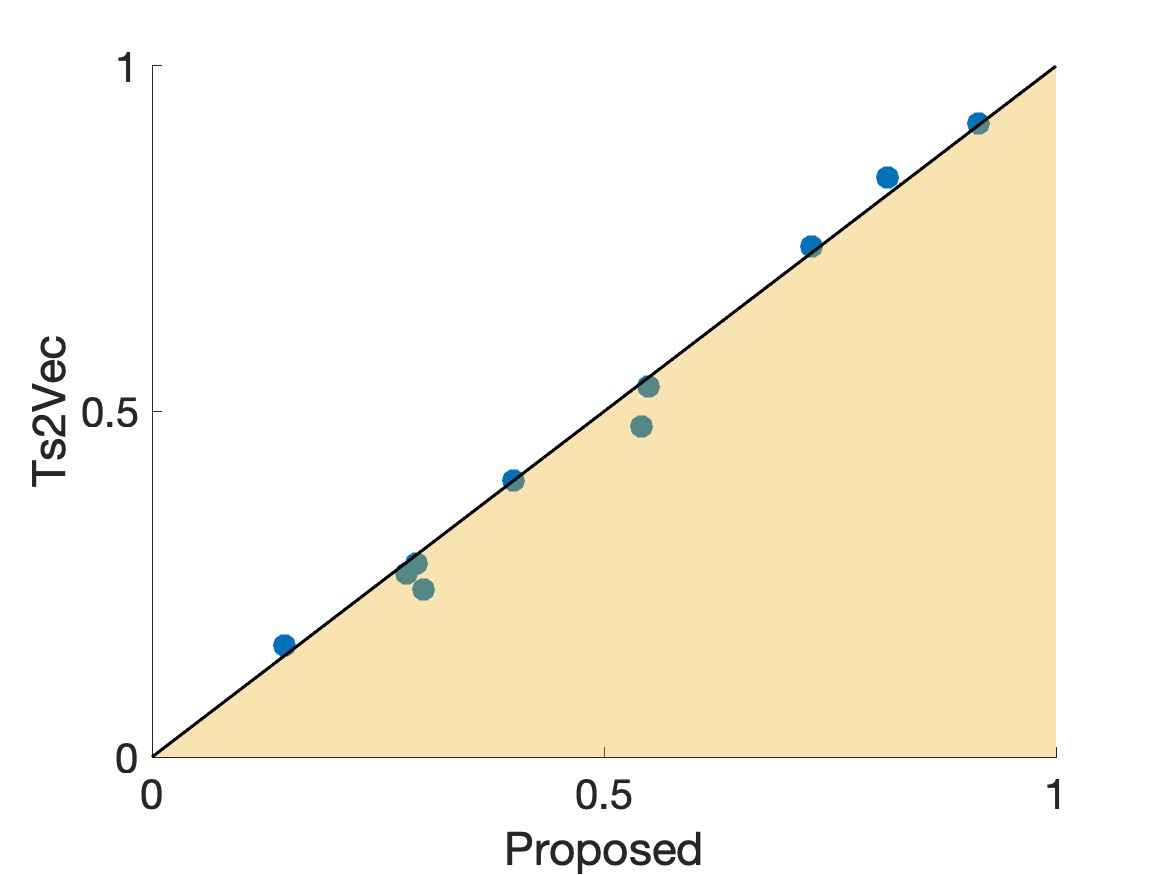}
    \caption{Accuracy ($K=16$)}
  \end{subfigure}
  \hfill
  \begin{subfigure}[t]{.23\textwidth}
    \centering
    \includegraphics[width=\linewidth]{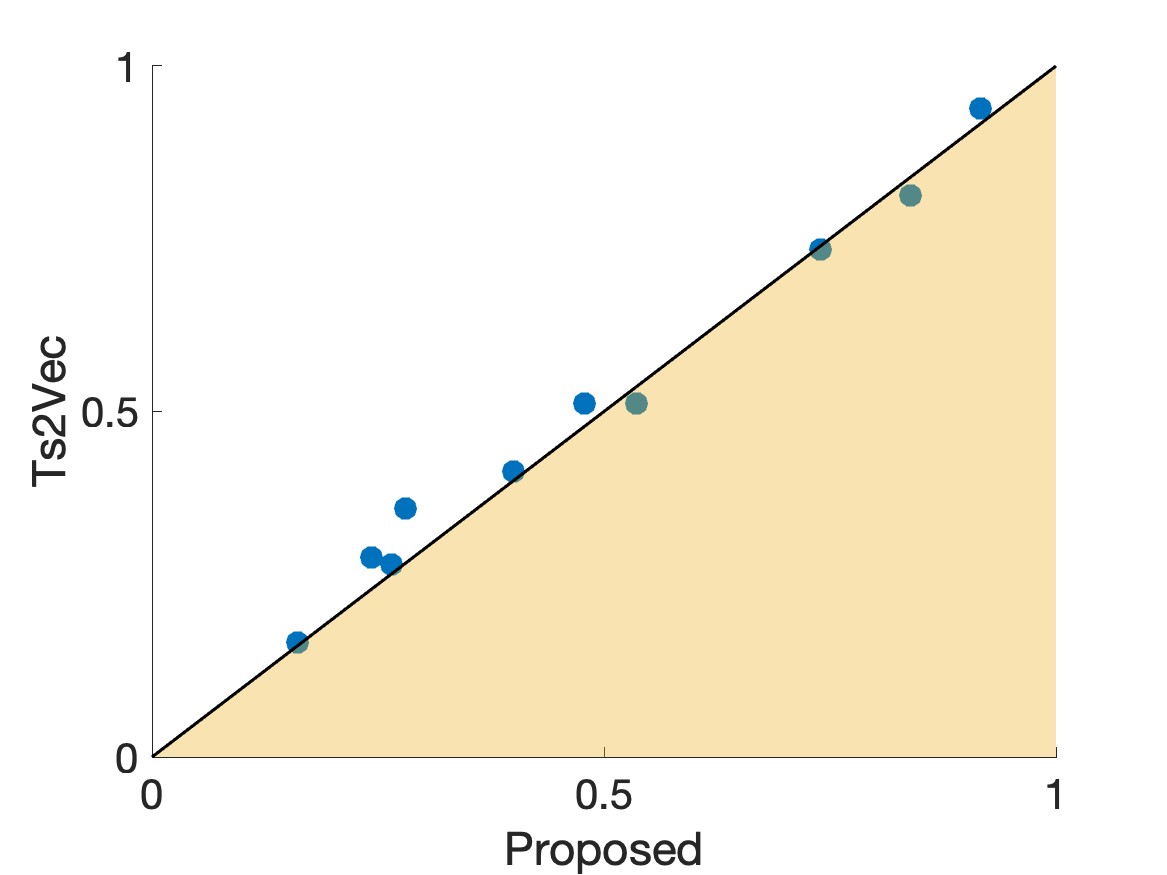}
     \caption{Accuracy ($K=32$)}
  \end{subfigure}

  \medskip

  \begin{subfigure}[t]{.23\textwidth}
    \centering
    \includegraphics[width=\linewidth]{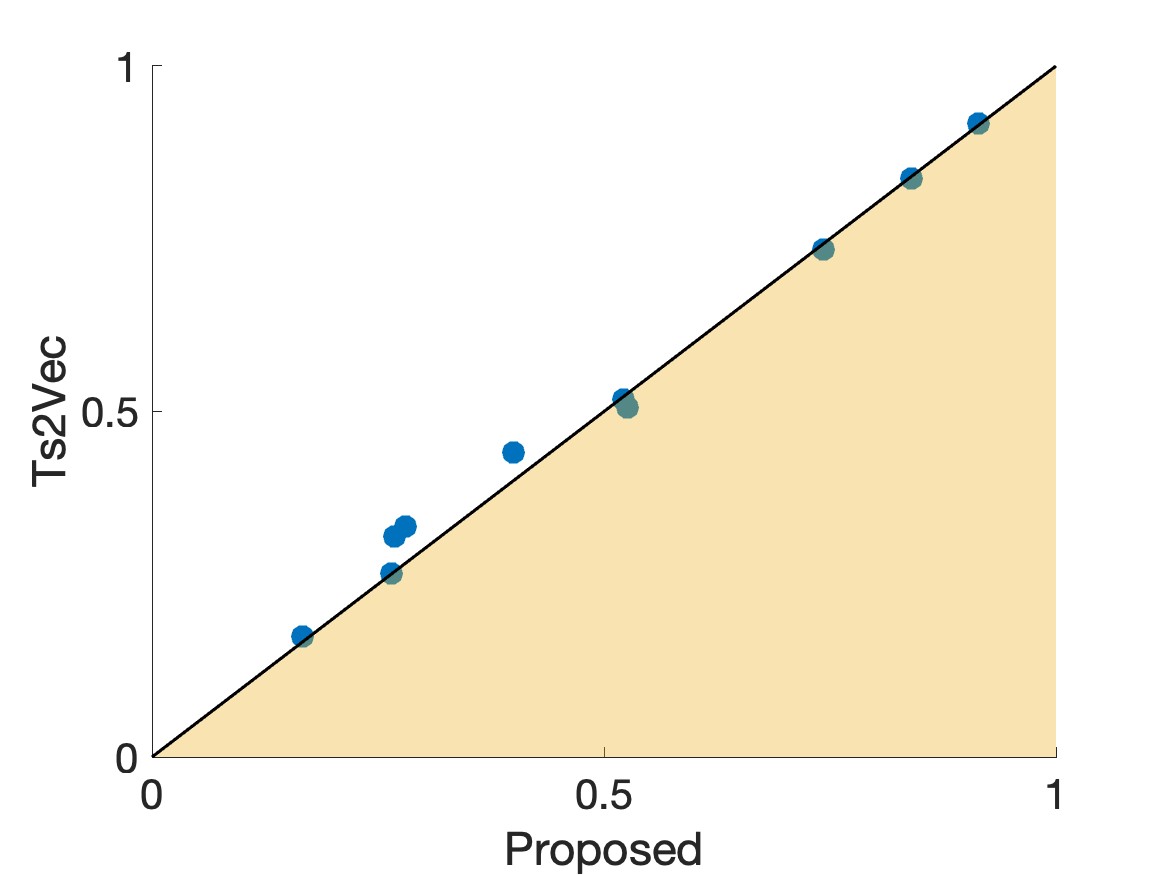}
    \caption{Accuracy ($K=64$)}
  \end{subfigure}
  \hfill
  \begin{subfigure}[t]{.23\textwidth}
    \centering
    \includegraphics[width=\linewidth]{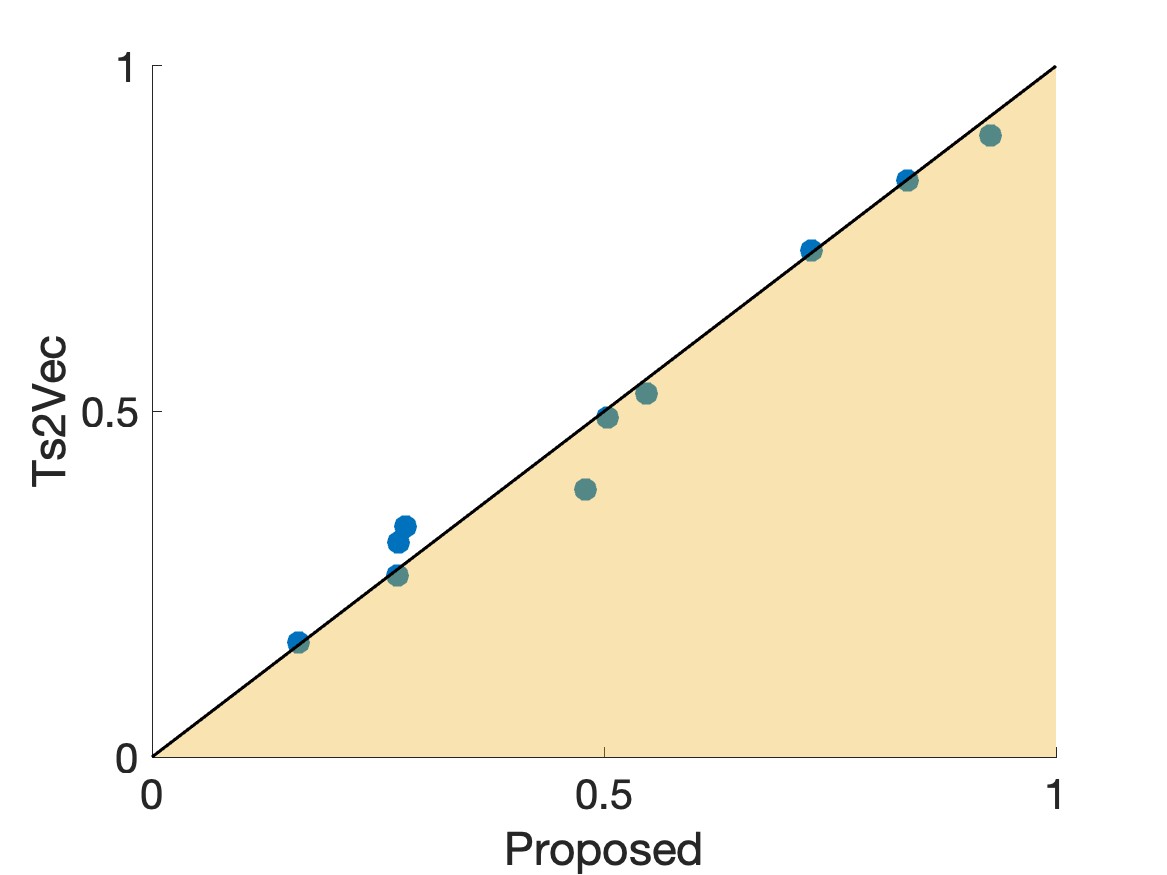}
     \caption{Accuracy ($K=128$)}
  \end{subfigure}
   \caption{Execution time and Accuracy comparison against Original Hierarchical Contrastive Learning Framework with Differing Embedding Size}
\end{figure}

\begin{table}[!h]
\centering
\resizebox{\columnwidth}{!}{%
\begin{tabular}{@{}lccccc@{}}
\toprule
                      & $\mathcal{L}^{hier}$        & $\mathcal{L}^{a}$         & 1-NN-DTW-D & 1-NN-DTW-I & 1-NN-ED \\ \midrule
HandMovementDirection &\textbf{ 0.311}           & 0.273          & 0.232      & \underline{0.306}      & 0.279   \\ \midrule
HeartBeat             &\textbf{0.733 }        & \underline{0.729}          & 0.717      & 0.659      & 0.659   \\ \midrule
AtrialFibrilation     & \textbf{0.333}          & \underline{0.280}          & 0.2        & 0.267      & 0.267   \\ \midrule
SelfRegulationSCP1    & \underline{0.835}          & \textbf{0.836  }         & 0.775      & 0.76       & 0.771   \\ \midrule
Phoneme               &\textbf{0.165}          & \underline{0.162}          & 0.151      & 0.151      & 0.151   \\ \midrule
SelfRegulationSCP2    & 0.526          & \textbf{0.547 }         & \underline{0.539}      & 0.533      & 0.483   \\ \midrule
Cricket               &  0.900         & 0.928          & \textbf{1 }         & \underline{0.986}      & 0.944   \\ \midrule
EthanolConcentration  & 0.263           &0.272        & \textbf{0.323 }     & \underline{0.304}      & 0.293   \\ \midrule
StandWalkJump         &  \underline{0.387}           & \textbf{0.480}          & 0.2        & 0.33       & 0.2     \\ \midrule
MotorImagery          & 0.492           & \underline{0.504}          & 0.5        & 0.39       & \textbf{0.51}    \\ \midrule
Average               & \underline{0.494} & \textbf{0.501} & 0.4637     & 0.4686     & 0.4557  \\ \bottomrule
\end{tabular}%
}
\caption{Non-SSL Baseline Comparisons (Accuracy Metric)}
\label{tab:my-table}
\end{table}

\begin{figure}
  \begin{subfigure}[t]{.23\textwidth}
    \centering
    \includegraphics[width=\linewidth]{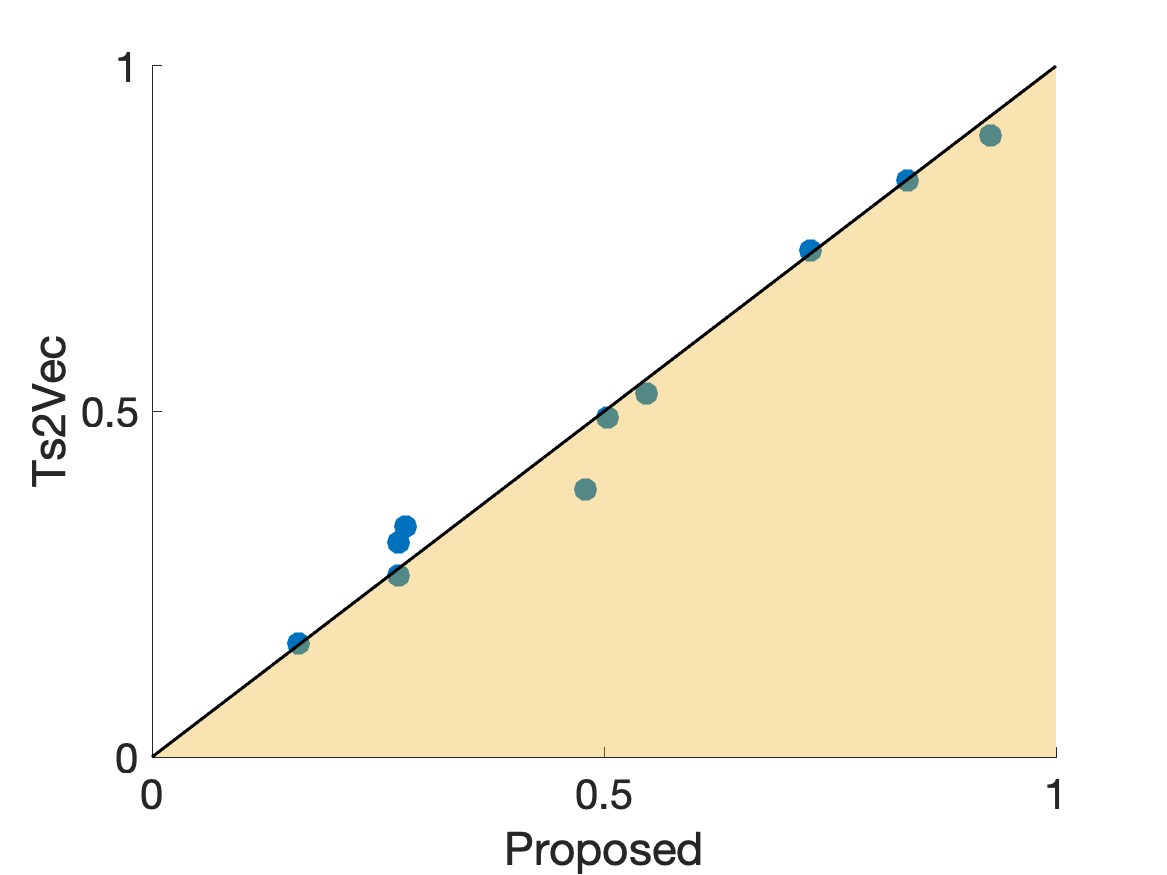}
    \caption{Proposed vs Original}
  \end{subfigure}
  \hfill
  \begin{subfigure}[t]{.23\textwidth}
    \centering
    \includegraphics[width=\linewidth]{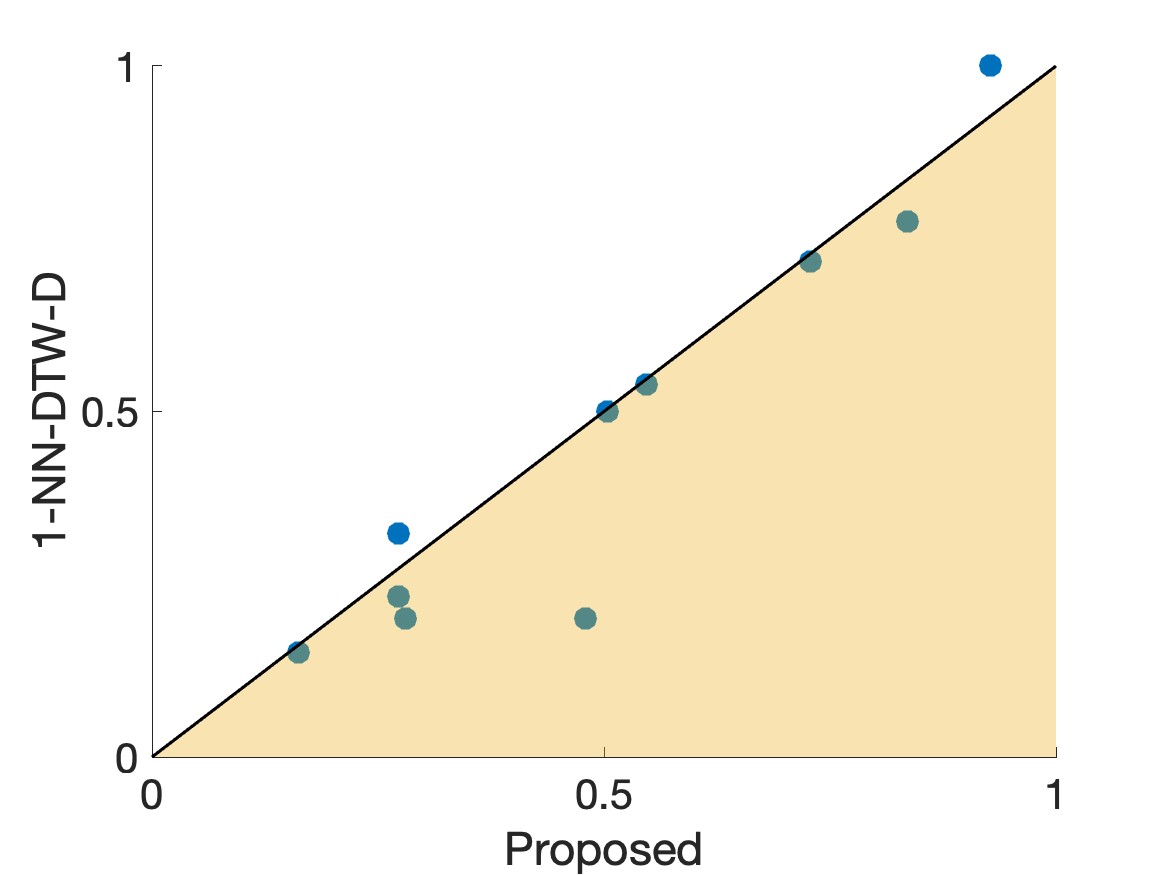}
    \caption{Proposed vs 1-NN-DTW-D}
  \end{subfigure}

  \medskip

  \begin{subfigure}[t]{.23\textwidth}
    \centering
    \includegraphics[width=\linewidth]{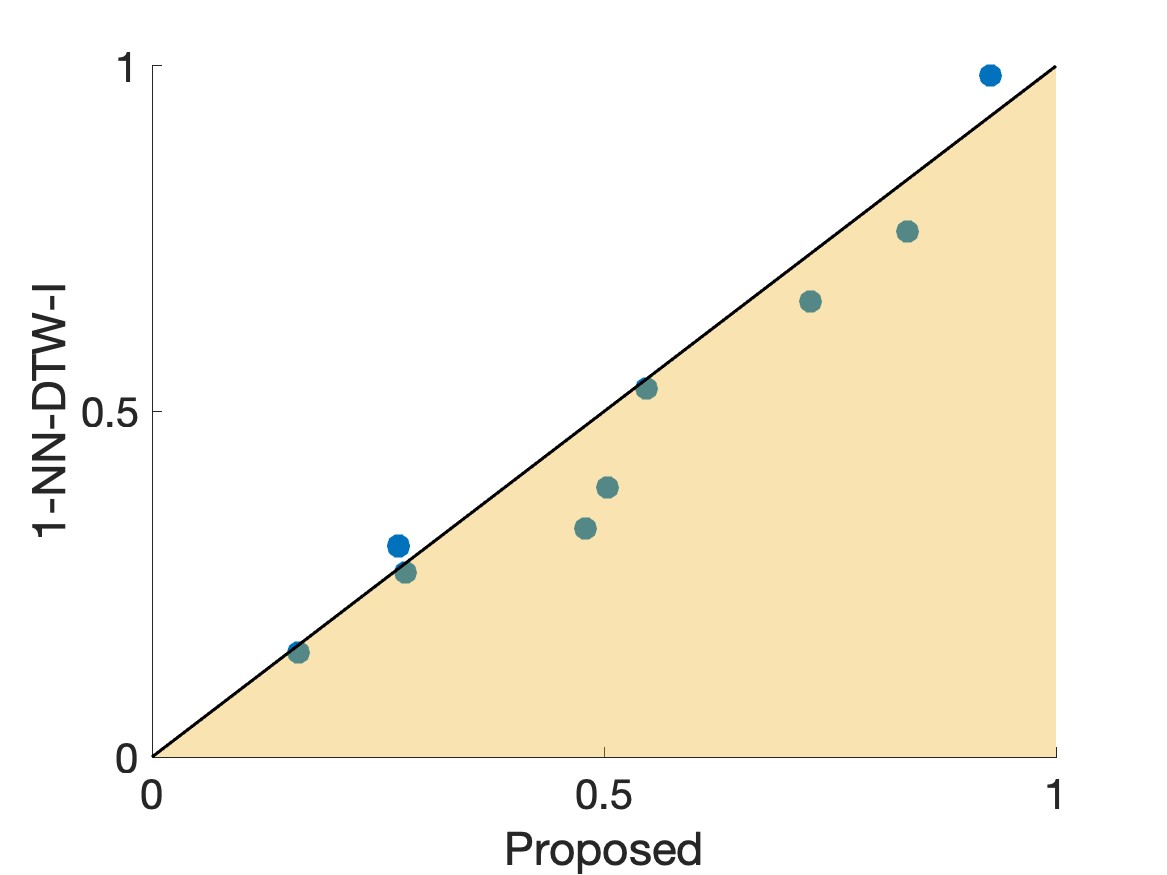}
    \caption{Proposed vs 1-NN-DTW-I}
  \end{subfigure}
  \hfill
  \begin{subfigure}[t]{.23\textwidth}
    \centering
    \includegraphics[width=\linewidth]{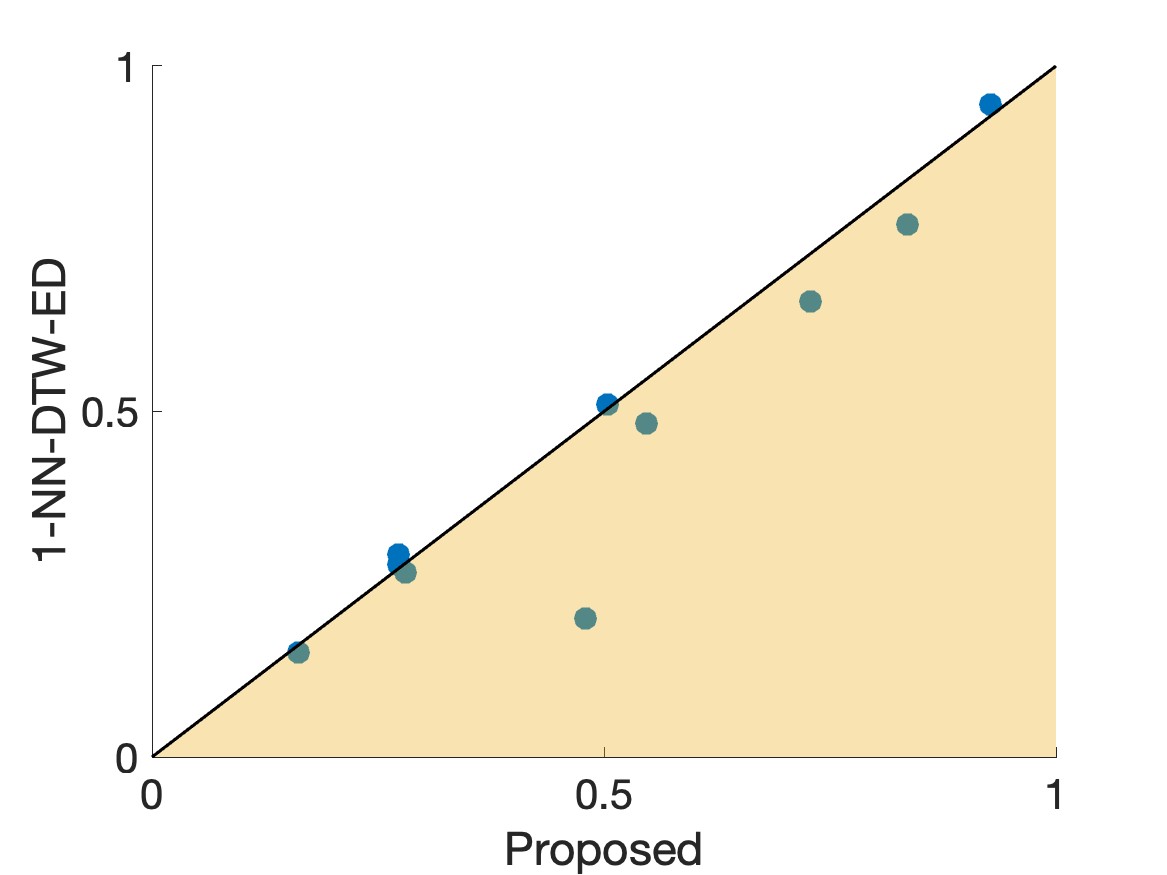}
    \caption{Proposed vs 1-NN-ED}
  \end{subfigure}
  \caption{Proposed Implementation vs Original and Non-SSL Baselines}
\end{figure}

\subsection{Vs. Non-SSL Baselines}

Next, the proposed method is compared with non-SSL baselines. The results are shown in Table III. Similar to Table II, the bold values indicate the best result and the underlining indicates the second best performance. 

According to the table, our approach achieves the highest average accuracy across all of the baselines. Additionally, in the table, the proposed contrastive learning framework consistently achieves top-1 and top-2 performance in most of the datasets. The proposed method vs. baselines one-to-one illustration is shown in Figure 6. From the result, we found that the proposed method consistently outperforms all of the selected baselines while achieving similar performance to the ts2vec's result. In summary, the results indicate that the proposed method's performance is similar to the original hierarchical contrastive learning framework and is better than the classical time series classification baselines.

\subsection{Efficiency Evaluation Across Time Series Length and Data Size}

\begin{table}[!h]
\centering
\resizebox{0.55\columnwidth}{!}{%
\begin{tabular}{@{}ccc@{}}
\toprule
Sample Length & $\mathcal{L}^{hier}$     & $\mathcal{L}^{a}$          \\ \midrule
100               & 30.12 & \textbf{24.36} \\ \midrule
200               & 38.66 & \textbf{30.45}  \\ \midrule
500               & 60.25 & \textbf{49.86} \\ \midrule
750               & 85.73  & \textbf{70.53} \\ \midrule
900               & 104.34 & \textbf{85.22} \\ \midrule
1000              & 116.15 & \textbf{95.11} \\ \midrule
3000              & 904.13 & \textbf{668.43} \\ \bottomrule
\end{tabular}%
}
\caption{Varying Sample Length Experiment in Seconds}
\label{tab:my-table}
\end{table}

In this subsection, we further evaluate the proposed method's efficiency compared with the original framework with  differing lengths and sizes for the time series dataset.

\subsubsection{Vs. Different Time Series Length}: For the scalability experiment for time series length, we increased the sample time series length from 100 to 3,000 and report the execution time of both frameworks. The parameters used for this experiment were as follows, batch size of 128 (32 for the last experiment), and an output embedding size of $K = 16$. The result is shown in Table IV. According to the table, we observed a faster execution time for our proposed method across all 7 trials compared to the original hierarchical learning framework. Moreover, we see that the longer the dataset is, the more significant the training time reduction is. The results show that our implementation is more efficient than the original method across differing time series lengths.

\begin{table}[!h]
\centering
\resizebox{0.55\columnwidth}{!}{%
\begin{tabular}{@{}ccc@{}}
\toprule
Sample Size & $\mathcal{L}^{hier}$     & $\mathcal{L}^{a}$            \\ \midrule
100               & 357.34 & \textbf{296.82} \\ \midrule
200               & 626.41 & \textbf{524.76} \\ \midrule
500               & 1469.21  & \textbf{1226.16} \\ \midrule
750               & 2227.72 & \textbf{1878.24} \\ \midrule
900               & 2699.86 & \textbf{2229.34} \\ \midrule
1000              & 2855.02 & \textbf{2386.30} \\ \bottomrule
\end{tabular}%
}
\caption{Varying Sample Size Experiment in seconds}
\label{tab:my-table}
\end{table}

\subsubsection{Vs. Different Sample Size}: For the second experiment, sample size, we opted for a varying sample size ranging from 100 to 1000, while maintaining a fixed time series length. The result is shown in Table V. According to the table, we observed that our proposed approach succeeded in outperforming the existing base model in all testing cases. Similiar to the time series length experiment, the sample size contributed to an increase in the execution time. Compared to the original model, we see a significant decrease in training time in larger sample sizes. 

In summary, the result indicates that the proposed method is much more efficient than the original framework across both time series length and data size.

\subsection{Embedding Visualization}

\begin{figure}
  \begin{subfigure}[t]{.23\textwidth}
    \centering
    \includegraphics[width=\linewidth]{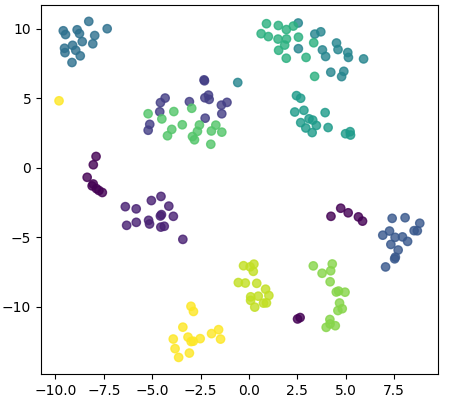}
    \caption{Cricket}
  \end{subfigure}
  \hfill
  \begin{subfigure}[t]{.23\textwidth}
    \centering
    \includegraphics[width=\linewidth]{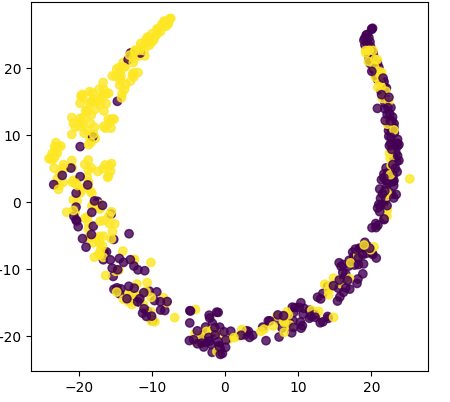}
     \caption{Self Regulation SCP1}
  \end{subfigure}

  \medskip

  \begin{subfigure}[t]{.23\textwidth}
    \centering
    \includegraphics[width=\linewidth]{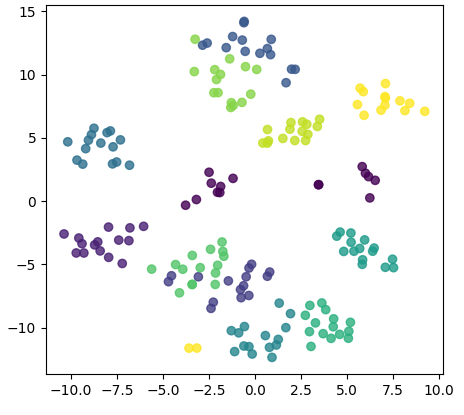}
    \caption{Cricket}
  \end{subfigure}
  \hfill
  \begin{subfigure}[t]{.23\textwidth}
    \centering
    \includegraphics[width=\linewidth]{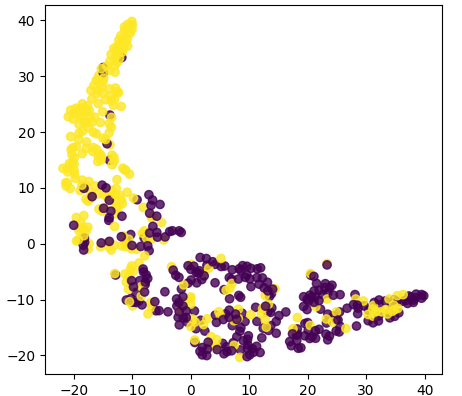}
     \caption{Self Regulation SCP1}
  \end{subfigure}
  \caption{t-SNE visualizations of Embedding Spaces: Proposed (top) vs Original (bottom)}
\end{figure}

We further demonstrate that the proposed method will produce similar representations compared with the original framework. In this experiment, we visualize the embedding vector obtained by both proposed framework and the original ts2vec framework via t-SNE visualization for Cricket and SCP1 datasets. The results are shown in Figure 7. The embedding visualizations obtained from original framework are visualizations c and d, and the embedding visualizations obtained via the proposed framework are visualizations a and b. We highlight the labels of each class with different colors for ease of understanding. Visually speaking, there are no significant differences in terms of representing the semantic meaning of the data. The result further demonstrates that the proposed framework can achieve similar embedding vectors compared with the original framework.

\section{Conclusion}

In this paper, we proposed a method to improve the computational efficiency of the ts2vec model for time series representation learning. Our method involves the use of an importance-aware resolution setting, in the model's loss function which allows us to reduce the computational load of the training process without sacrificing the model's performance in time series classification. Our experimental results confirm our proposed method is effective on both small and large scale datasets. Our model achieved similar classification accuracy in a range of 10 UCR/UEA datasets while consistently reducing the training time. These findings suggest that our method can be a valuable tool for researchers and practitioners working with large-scale time series datasets.

\section{Acknowledgment}

This work is supported by the NSF under Grant CNS-2318682, IIS-2348480, and U.S. Department of Education under GAANN (Grant No. 5100001128). 

\bibliographystyle{plain}
\bibliography{sample}

\end{document}